\documentclass{article}
\usepackage{spconf,amsmath,graphicx}
\usepackage{multirow}
\usepackage{mathtools}
\usepackage[normalem]{ulem}
\usepackage{booktabs}
\usepackage{xcolor}
\usepackage{hyperref}

\usepackage{soul, color}
\makeatletter
\def\SOUL@geheimunderline#1{{%
    \setbox\z@\hbox{#1}%
    \dimen@=\wd\z@
    \dimen@i=\SOUL@uloverlap
    \advance\dimen@2\dimen@i
    \rlap{%
        \null
        \kern-\dimen@i
        \SOUL@ulcolor{\SOUL@ulleaders\hskip\dimen@}%
    }%
    \phantom{#1}%
}}
\DeclareRobustCommand{\secret}{%
  \let\SOUL@ulunderline\SOUL@geheimunderline
  \SOUL@hlsetup\SOUL@}
\makeatother

\title{The Relevance of AWS Chronos: An Evaluation of Standard Methods for Time Series Forecasting with Limited Tuning}
\twoauthors
 {Matthew Baron\sthanks{matthew.baron@nike.com}}
	{\textit{NIKE}}
 {Alex Karpinski\sthanks{alexkarp@umd.edu}}
	{University of Maryland, College Park}

\begin{document}
\pagenumbering{arabic}
\sethlcolor{gray}
\maketitle
\begin{abstract}
A systematic comparison of Chronos, a transformer-based time series forecasting framework, against traditional approaches including ARIMA and Prophet. We evaluate these models across multiple time horizons and user categories, with a focus on the impact of historical context length. Our analysis reveals that while Chronos demonstrates superior performance for longer-term predictions and maintains accuracy with increased context, traditional models show significant degradation as context length increases. We find that prediction quality varies systematically between user classes, suggesting that underlying behavior patterns always influence model performance. This study provides a case for deploying Chronos in real-world applications where limited model tuning is feasible, especially in scenarios requiring longer prediction.
\end{abstract}
\section{Introduction}
\label{sec:intro}
This study explores the relevance and applicability of AWS Chronos, a powerful tool for time series forecasting. By comparing Chronos to standard industry methods, the research evaluated the effectiveness of established techniques when applied to time series forecasting tasks, particularly when limited tuning or optimization is permitted. The study used a single dataset comprising bicycle rental records from the Capital Bike-share program in Washington, D.C., as representative of demand based signals commonly seen in industry contexts. The study examined the suitability and potential advantages of AWS Chronos in specific forecasting scenarios, offering insights into when and how this tool can be best utilized within the broader context of industry-standard practices.

\section{Methods}
\label{sec:methods}
This section describes the data collection and preparation stage, time series models used in the research, and complete procedures for model building and evaluation. 

\subsection{Data Description and Preparation}
\label{ssec:data}

Our evaluation focuses on a single dataset comprising bicycle rental records from the Capital Bike-share program in Washington, D.C., spanning 2011 to 2012. This dataset is publicly known as the UCI Bike Sharing \cite{fanaee-tBikeSharing2013} dataset and can be accessed at \href{https://doi.org/10.24432/C5W894}{[10.24432/C5W894]}. Each data point in the original dataset represents an individual bike rental, including a timestamp accurate to the minute along with various additional attributes.
To prepare the dataset for analysis using Chronos, we extracted only the timestamp and a binary indicator of customer type (casual or registered). Registered customers pay a subscription fee for unlimited rentals, while casual customers encompass all other users.
To ensure uniform intervals between data points, as required by our models, we aggregated the data to a daily basis, summing the total number of bikes rented per day. The source code for this data transformation and aggregation process is available at \cite{lxkarpLxkarpTimeseriesmodeling2024}.

As the dataset included the indicator of a casual versus registered user, we decided to use this to further divide the subsets into two equal-length single time-series, resulting in six distinct evaluation subsets.  Rather than sum the series together to predict the total usage of the bike-share, we hypothesize that any customer's individual usage of the bike-share program would differ significantly based on type of user. These different types of customers have a different psychological relationship to their preferences for when and how often to rent a bike. Effectively, the two categories represent different products that are offered by the bike-share program. We wanted to evaluate the performance of each of our chosen models given that each model itself represents a different interpretation on how time-series behave and move forward.

We evaluated the performance of 4 models on six prediction targets, created by combining two user categories (casual and registered) with three different prediction periods (Week10, July, and Q4).

\begin{itemize}
    \setlength\itemsep{0.2em}
    \item Short-term (Week 10): \textbf{n} 7-day weeks context, 1 week prediction
    \item Medium-term (July): \textbf{n} 31-day months context, 1 calendar month prediction
    \item Long-term (Q4): \textbf{n} 91-day quarters, 1 91-day quarter prediction
\end{itemize}

For each prediction target, we tested four different context-to-prediction ratios (2:1, 3:1, 4:1, and 5:1), allowing us to evaluate how the amount of historical context affects prediction quality. Combined with our two user types (casual and registered), this created 24 distinct evaluation scenarios (2 user types × 3 prediction targets × 4 context ratios).

\begin{table}[ht]
\caption{Evaluation Structure}
\label{tab:eval-structure}
\begin{tabular}{ll}
\toprule
Component & Variants \\
\midrule
User Types & Casual, Registered \\
Predictions & Week10 (7d), July (1mo), Q4 (3mo) \\
Context Ratios & 2:1, 3:1, 4:1, 5:1 \\
Models & ARIMA, Prophet, Chronos, Naive \\
\midrule
Total Scenarios & 24 (2×3×4) \\
Total Evaluations & 96 (24 scenarios × 4 models) \\
\bottomrule
\end{tabular}
\end{table}

\subsection{Framework}
\label{ssec:framework}
For each of the 24 evaluation scenarios (Week10, July, Q4) $\bigtimes$ (Casual users, Registered users) $\bigtimes$ (2:1, 3:1, 4:1, 5:1) individual models were constructed using (auto)ARIMA, Prophet, Chronos, and a seasonal-naive baseline. We recognized the complexities in optimizing/tuning our models to address accuracy and robustness for our particular dataset. Therefore, we chose to analyze our model/data pairing with minimal tuning of the model, representative of an analyst in industry doing a POC to decide where to invest further efforts. 

\subsection{Background of Chronos}
\label{ssec:chronos}
Chronos \cite{ansariChronosLearningLanguage2024}, a novel framework designed for probabilistic time series forecasting, leverages existing large language model (LLM) architectures with minimal adaptations. 
The core innovation of Chronos lies in its tokenization process, where time series values are transformed into discrete tokens through scaling and quantization.
This tokenization allows the model to utilize a fixed vocabulary, enabling it to train on standard transformer-based architectures without the need for time-series-specific modifications. By employing a categorical distribution to model observations, Chronos effectively performs regression via classification. 
This Chronos approach diverges from traditional probabilistic forecasting methods that rely on continuous distributions (such as Quantile Regression Averaging) and generative or representational forecasting models (such as ARIMA or Prophet) that require fitting to observations of the random process.

\begin{figure}[h]
    \centering
    \includegraphics[width=1\linewidth]{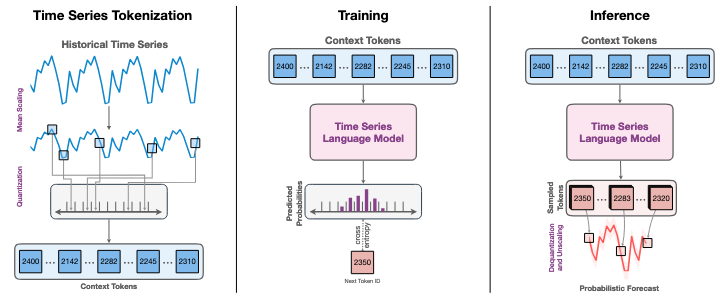}
    \caption{High-level depiction of Chronos: Transform input time series into tokens via scaling an quantization; train the language model via cross-entropy loss; autoregressively sample multiple trajectories, then map to numerical values to obtain predictive distribution.}
    \label{fig:enter-label}
\end{figure}

The training of Chronos models involves pre-training on a diverse collection of datasets, including both real and synthetic data generated through Gaussian processes. This extensive training should enhance the model's generalization capabilities, allowing remarkable zero-shot performance on unseen datasets. In a benchmark involving 42 datasets, Ansari et. al demonstrated superior performance compared to classical local models and task-specific deep learning methods, excelling in in-domain tasks while also showing competitive results on new datasets. The design of Chronos as a framework requires minimal changes to existing LLM architectures, positioning it as a versatile pre-trained tool for various time series forecasting applications, which may be easily updated. Chronos is designed to be an off-the-shelf solution for any industry task. 

\subsection{Background of Prophet}
\label{ssec:prophet}
The Prophet algorithm is designed to handle the complexities of time series forecasting, particularly in business contexts where data may exhibit various seasonal patterns and trends. At its core, Prophet frames the forecasting problem as a curve-fitting exercise, which differs from traditional time series models that focus on temporal dependencies in the data. The model incorporates three main components: a trend function \(g(t)\) that captures non-periodic changes, a seasonal component \(s(t)\) that accounts for periodic fluctuations (such as weekly and yearly seasonality), and a holiday effect \(h(t)\) that models the impact of holidays, which can occur irregularly.

$$y(t)=g(t)+s(t)+h(t)+\epsilon_t$$

One of the key advantages of Prophet is its ability to work with irregularly spaced data, eliminating the need for interpolation of missing values, a common requirement in ARIMA models. This flexibility allows analysts to quickly fit the model and explore various specifications interactively. The model's parameters are designed to be intuitive, enabling analysts with domain knowledge to adjust them without needing deep expertise in the underlying statistical methods. Although Prophet sacrifices some inference benefits associated with purely generative models like ARIMA, it generates a statistical distribution of uncertainty in future forecasts. This uncertainty can be useful as an evaluation signal of forecast quality for analysts. \cite{taylorForecastingScale2017}

\subsection{Background of ARIMA}
\label{ssec:arima}
ARIMA (AutoRegressive Integrated Moving Average) models are a class of statistical models used for analyzing and forecasting time series data. These models combine three components: autoregression (AR), integration, (I), and moving average (MA) The AR component captures the relationship between an observation and a certain number of lagged observations, functioning as a self-similarity measure. The I component represents differencing of raw observations to achieve stationarity, specifically removal of non-constant trends. This is roughly equivalent to a high-pass filter of the time-series in order to remove long-time/non-constant variation. The MA component incorporates the dependency between an observation and a residual error from a moving average model applied to lagged observations. A representation of how far the current observation is from the expectation given a historical window of observations. Formally an ARIMA(\(p,d,q\)) model representing time series data \(X_t\) indexed by \(t\) is expressed as:

$$
\left( 1 - \sum_{i=1}^p \varphi_i L^i \right)
(1-L)^d X_t
= \left( 1 + \sum_{i=1}^q \theta_i L^i \right) \varepsilon_t \,
$$

Where $L$ is the lag operator, the $\varphi_i$ are parameters for the AR model, $\theta_i$ are parameters for the MA, and $\varepsilon_t$ are the error terms. 
The models parameters, (\(p,d,q\)) determine the order of the AR, I, and MA parts respectively. Estimation of these parameters typically involves methods such as maximum likelihood estimation, or conditional least squares, often utilizing either Akaike or Bayesian Information Criteria to determine the model's fidelity to reality and perfect fit respectively.  The autoARIMA process is used to identify the most optimal orders for (\(p,d,q\)) and settles on a single, fit ARIMA model which is used for prediction.

\subsection{Performance Evaluation}
\label{ssec:metrics}
In this study, three evaluation metrics were used to assess the models' performance; Earth-Mover Distance (EMD), Mean Absolute Square Error (MASE), and Weighted Quantile Loss (WQL) to comprehensively assess and compare the performance of our models. We used both WQL and MASE to maintain consistency with the original evaluations as described in the Chronos paper \cite{ansariChronosLearningLanguage2024}. The Chronos authors describe wanting to assess both the probabilistic and point forecast performance, and chose WQL \cite{amazonforecastEvaluatingPredictorAccuracy} and MASE to represent each respectively. 
WQL represents an assessment of the probabilistic forecast, because of its relation to the continuous ranked probability score (CRPS). The reported WQL is calculated by sampling the probability distribution at the decimal quantiles \{0.1, 0.2, ... , 0.9\}, then averaging across them. We decided that in addition to the WQL as a representation of the probabilistic forecast performance, we would calculate the Earth-Mover's Distance (EMD) \cite{baronSuitabilityEarthMoverDistance2025} between the actuals and each of the decimal quantiles, similarly taking the mean across quantiles. Because the forecasts represent bicycle users at points in time, each sample is a density (of rides) over the metric space (days). This cost value is then a proxy for the actual fiscal cost of the probabilistic error from the forecast. 

\section{Experiments}
\label{sec:experiments}

\subsection{Process Description}
\label{ssec:process}
Our evaluation process consists of two main stages:

\begin{enumerate}
    \item Forecasting
    \item Visualizing \& Summarizing
\end{enumerate}

While our forecasting methodology aligns with Amazon's "zero-shot" evaluation approach, our paper diverges in the visualization and summarization techniques employed. The analysis provided by Amazon comes in the form of metrics (WQL and WMAPE) which aggregate the results of forecasting 27 datasets.

\subsection{Benchmark/Baseline Performance}
\label{ssec:benchmark}
Our forecasting process closely follows the methods described in the "zero-shot" evaluation section of the Amazon paper. For Chronos, we selected the \texttt{'chronos-t5-small'} model and configured our test harness to use the same hyper-parameters as Amazon, specifically for \texttt{'batch\_size'} and \texttt{'num\_samples'}, largely using the gluonTS interface to Chronos \cite{gluonts_arxiv}. The context data as described above was provided equally to all models.

We performed forecasts using each of our four models (Chronos, Prophet, AutoARIMA, and Naive) against our six prediction targets. For each target, we tested four different context-to-prediction ratios (2:1, 3:1, 4:1, and 5:1), resulting in a total of 24 evaluation scenarios (6 targets × 4 ratios). This design allows us to evaluate not only how models perform on different prediction windows and user types, but also how sensitive each model is to the amount of historical context provided.

The forecasting test-bench is made available alongside our data preparation code in our repository. Additionally, graphs of the raw predictions from each model, category, segment, and context ratio are stored there for inspection \cite{lxkarpLxkarpTimeseriesmodeling2024}.

\section{Results}
\label{sec:results}

\begin{figure*}
    \centering
    \includegraphics[width=1\linewidth]{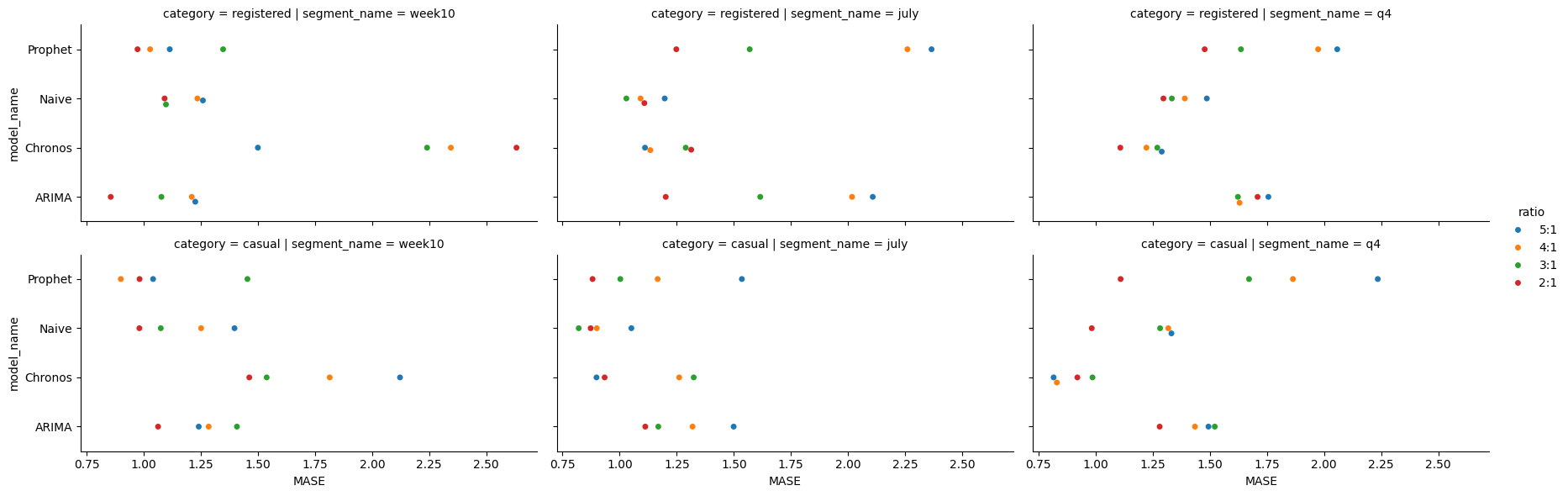}
    \caption{Model Performance Comparison (MASE)}
    \label{fig:MASE-swarm}
\end{figure*}

\begin{figure*}
    \centering
    \includegraphics[width=1\linewidth]{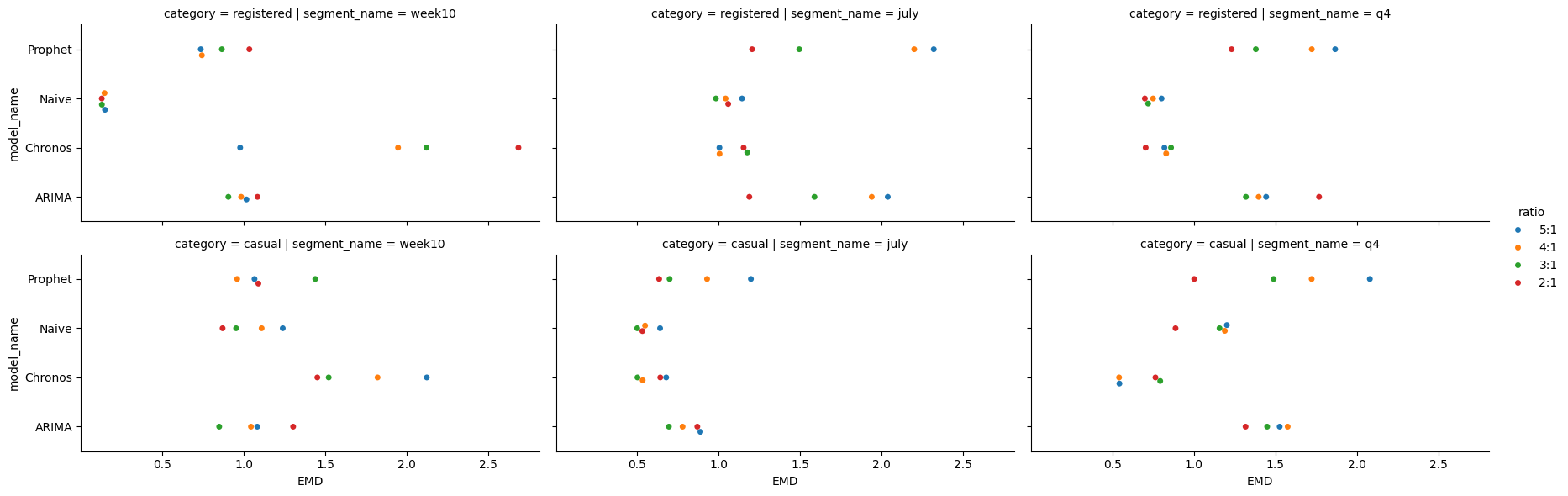}
    \caption{Model Performance Comparison (EMD)}
    \label{fig:EMD-swarm}
\end{figure*}

\begin{figure*}
    \centering
    \includegraphics[width=1\linewidth]{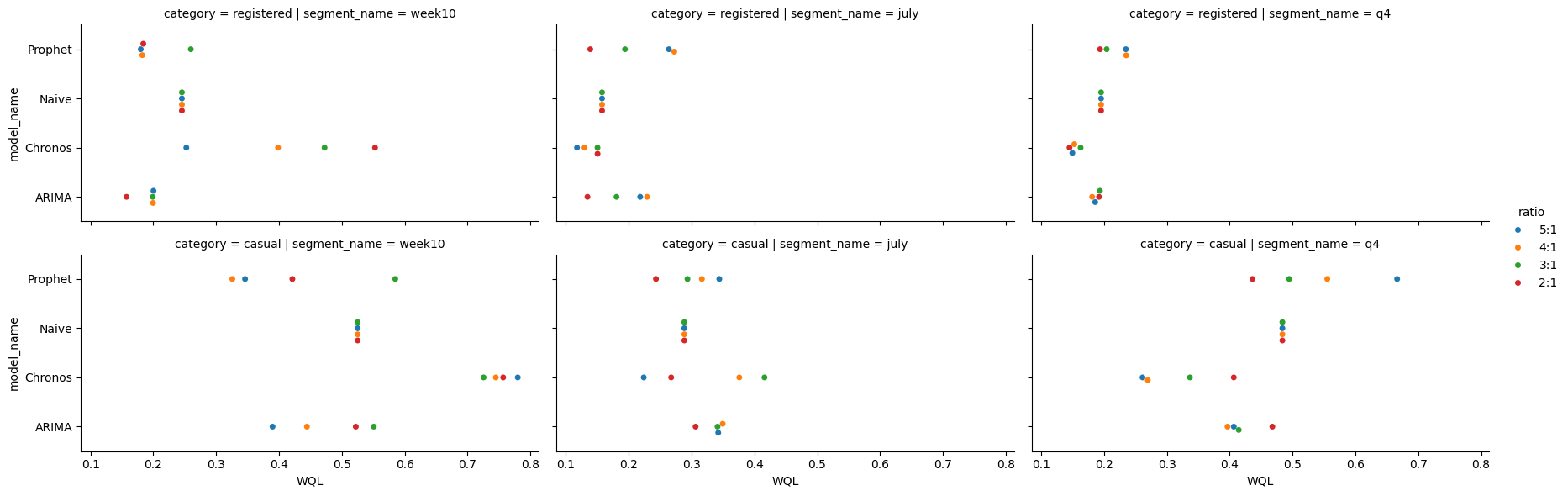}
    \caption{Model Performance Comparison (WQL)}
    \label{fig:WQL-swarm}
\end{figure*}

Our results are presented through two complementary formats:
\begin{itemize}
    \item Swarm plots (Figures \ref{fig:MASE-swarm}, \ref{fig:EMD-swarm}, \ref{fig:WQL-swarm}) providing visualization of model performance across all evaluation scenarios
    \item Detailed metric tables (Tables \ref{tab:wql-degradation-july}-\ref{tab:wql-degradation-q4}) showing precise percentage changes in performance as context length increases
\end{itemize}

\subsection{Reading the Swarm Plots}
\label{ssec:swarm-plots}
Each swarm plot is organized into panels by prediction target (Week10, July, Q4) and user type (Casual, Registered). Within each panel, models are shown on the vertical axis, with their performance scores distributed horizontally. The color of each point in the swarm corresponds to the context-prediction ratio. The x-axis scale is consistent across all plots within a metric, facilitating a direct comparison of the model scores. 

The tables \ref{tab:calendar}, \ref{tab:two-to-one}, \ref{tab:three-to-one}, \ref{tab:four-to-one}, \ref{tab:five-to-one}  present a comprehensive evaluation of model performance, organized by the different analysis ratios. Calendar Split Table \ref{tab:calendar} is the initial work, where the data is split on calendar boundaries i.e. the context for "july" contains April 1 to June 30, rather than March 30 to June 30 in the 3:1 ratio data. They offer a detailed comparison of our metrics for each model across the various segments and categories. We've bolded the lowest EMD, italisized the lowest WQL and underlined the lowest MASE in each column for clarity. 

\section{Discussion}
\label{sec:discussion}

Evaluation of our results reveals several noteworthy patterns in the performance of our selected timeseries forecasting models, particularly in their application to different product categories and prediction horizons. We organize our discussion around three key findings:
\begin{enumerate}
    \item the differential performance between user types
    \item the performance of models across prediction horizons
    \item the impact of context length on forecast quality.
\end{enumerate}

\subsection{Differential Performance Patterns in User Types}
\label{ssec:behavior}
Across all models and time horizons, predictions for registered users achieved lower error metrics compared to casual users.

The dot swarm charts show a consistent clustering of registered dots (top row) to the left of their casual counterparts (bottom rows).

This systematic difference in prediction quality suggests that registered users exhibit more predictable behavior patterns, maybe due to routine usage such as regular commuting. The relative stability of registered user predictions is especially evident in the lower WQL scores.

\subsection{Model Performance Across Prediction Horizons}
\label{ssec:pred-horizons}
Chronos performed especially well in longer prediction tasks. Despite the model's documented 64-token limitation for prediction length, Chronos demonstrated strong performance in Q4 predictions, achieving the lowest WQL scores among all models for both user types. This contradicts initial expectations and suggests that the model's effective range may extend beyond its stated limitations for certain types of time series.

The performance pattern across different prediction horizons reveals distinct strengths for different models. The seasonal naive model shows remarkable effectiveness for short-term predictions, particularly evident in Week10 forecasts where it achieved winning or nearly-winning low scores across all of our error measurements. This suggests that for short-term bike-sharing predictions, the assumption that patterns will repeat from the previous week may have some unexpected predictive power.

Conversely, Prophet consistently underperformed compared to Chronos and ARIMA across all timeframes, with particularly high error scores in Q4 predictions. This suggests that either the time series of interest doesn't conform well to Prophet's model, or that Prophet wasn't given enough exogenous information. The data demonstrates both a visible trend component and micro and macro seasonality, specifically periodicity at the weekly granularity as well as the annual cycle reflective of the seasons. All models in this analysis were only given the time-series itself to rely on for training and prediction. It is possible that vanilla Prophet does not have the holiday signal baked in, and may be reliant on users providing such a signal as an exogenous variable. An additional possibility is that Prophet itself is rather reliant on the presence of additional time-series in order to generate reliable and accurate predictions.

\subsection{Impact of Context Length}
\label{ssec:context-impact}
Our analysis of varying context-to-prediction ratios revealed a familiar problem: increasing the context length does not necessarily improve forecast accuracy. While all models show some sensitivity to context length, the pattern and magnitude of this effect varied significantly between our selected models.

Both ARIMA and Prophet demonstrate clear degradation as context length increases, particularly for registered users in July predictions. ARIMA's performance deteriorates substantially, showing a +34.4\% increase in WQL at 3:1 ratio, worsening to +70.6\% at 4:1, and remaining significantly degraded at +62.4\% with a 5:1 ratio. Prophet shows even more dramatic sensitivity to context length, with its WQL scores deteriorating progressively from +39.8\% at 3:1 to +96.1\% at 4:1, maintaining this severe degradation (+90.0\%) at 5:1. This degradation aligns with known limitations of traditional forecasting approaches, where longer context lengths can amplify noise, increase susceptibility to concept drift, and lead to overfitting.

\begin{table*}[h]
\caption{July: Percentage Change in WQL Scores from 2:1 Ratio Baseline}
\label{tab:wql-degradation-july}
\centering
\begin{tabular*}{0.6\textwidth}{@{\extracolsep{\fill}} l|rrr|rrr}
\toprule
 & \multicolumn{3}{c|}{Casual} & \multicolumn{3}{c}{Registered} \\
Model & 3:1 & 4:1 & 5:1 & 3:1 & 4:1 & 5:1 \\
\midrule
ARIMA   &  +11.4\% &  +14.1\% &  +11.8\% &  +34.4\% &  +70.6\% &  +62.4\% \\
Prophet &  +20.6\% &  +30.0\% &  +41.4\% &  +39.8\% &  +96.1\% &  +90.0\% \\
Chronos &  +55.5\% &  +40.5\% &  -16.3\% &   -0.1\% &  -13.8\% &  -21.7\% \\
\bottomrule
\end{tabular*}
\end{table*}

This degradation with increased context length may be attributed to a failure of the model itself: longer context windows may introduce noise and seasonal variations that push the model's ability to identify relevant patterns. It's also possible that the bike-sharing system itself underwent change, making distant historical data irrelevant for future predictions.

Chronos, however, shows a markedly different pattern. For registered users in July, Chronos maintains relatively stable performance, showing slight improvements as context increases (-0.1\% at 3:1, -13.8\% at 4:1, and -21.7\% at 5:1). Even more striking is its behavior with Q4 casual predictions, where it shows consistent improvement in WQL scores (-17.2\%, -33.6\%, and -35.7\% for 3:1, 4:1, and 5:1 ratios respectively).

This differential response to increased context length suggests that Chronos's architecture provides resistance to traditional overfitting problems. While both classical time-series models like ARIMA and decomposition-based approaches like Prophet can become overwhelmed by additional historical data, Chronos appears to maintain or even improve its predictive power. This resistance to context-related degradation represents a significant advantage in real-world applications where abundant historical data is available.

\begin{table*}[ht]
\caption{Q4: Percentage Change in WQL Scores from 2:1 Ratio Baseline}
\label{tab:wql-degradation-q4}
\centering
\begin{tabular*}{0.6\textwidth}{@{\extracolsep{\fill}} l|rrr|rrr}
\toprule
 & \multicolumn{3}{c|}{Casual} & \multicolumn{3}{c}{Registered} \\
Model & 3:1 & 4:1 & 5:1 & 3:1 & 4:1 & 5:1 \\
\midrule
ARIMA   &  -11.4\% &  -15.3\% &  -13.1\% &   +0.7\% &   -5.7\% &   -3.2\% \\
Prophet &  +13.4\% &  +27.3\% &  +52.8\% &   +5.4\% &  +21.5\% &  +21.2\% \\
Chronos &  -17.2\% &  -33.6\% &  -35.7\% &  +12.2\% &   +5.3\% &   +3.1\% \\
\bottomrule
\end{tabular*}
\end{table*}

\subsection{Relation to Prior Work}
\label{ssec:priorwork}

Simple Combination of Univariate Models\cite{petropoulosSimpleCombinationUnivariate2020} defines the SCUM ensemble as the median combination of the point forecasts of ETS, CES, ARIMA, and Dynamic Optimized Theta. Since proposed in 2020, SCUM has been a reference of performance for many new time series prediction projects. 

As cited in the original Chronos paper \cite{ansariChronosLearningLanguage2024}, specifically Benchmark II, the Chronos model demonstrates superior performance to the SCUM combination of models by a slim margin of Agg. Relative WQL. However, it is important to note that SCUM outperformed Chronos' smaller variants (\texttt{small} and \texttt{mini}) in Agg. Relative MASE. 

A commonly cited comparison in discussions about Chronos performance is Nixtla's comparison\cite{nixtlaNixtlaExperimentsAmazonchronos}. The write-up, although lacking in rigorous discussion, presents an interesting comparison. On a selection of monthly, quarterly, and yearly datasets, the SCUM combination outperformed Chronos by 10\% across accuracy metrics (CRPS, MASE, and SMAPE). Additionally, SCUM achieved this with only 20\% of the wall time, indicating a significant efficiency advantage. 

The LinkedIn post \cite{ansariabdulfatirExtendedComparisonChronos} by A. F. Ansari, a Chronos author, provides a valuable rebuttal to Nixtla's comparison, commenting on the model's development and limitations. Ansari acknowledges that the Nixtla selection of datasets was biased towards short, low-frequency time series, which may have influenced the model's relative performance characteristics, offering to extent the benchmark of the paper in a pull request\cite{oleksandrshchurExtendChronosEvaluation}.  Chronos was trained on only 2 of the 28 datasets indexed at a monthly frequency, with none at a lower frequency.  Benchmark II, contained a more diverse set of datasets, with 7 out of 27 indexed less frequently than monthly and an additional 6 indexed monthly. This disparity in training data may contribute to the observed performance differences when tested with lower frequency data. He also suggests the decision to use such an ensemble was made in hindsight, or with humans in the loop. This criticism rings hollow as Benchmark II contains the SCUM ensemble. 

Although Nixtla highlights Chronos' wall time deficiency, Ansari calls out the comparably equal computational costs especially for smaller variants, with minimal trade-off in accuracy. He also mentions TimeGPT, an earlier proprietary model, as a deprecated option, suggesting that Chronos is the preferred choice. 

In a separate LinkedIn post\cite{desaiNewBenchmarkingStudy2024}, A. Desai of the Ready Tensor project provides an additional extensive evaluation of Chronos, ranking it highly for zero-shot forecasting accuracy, on par with full-shot machine-learning and neural network models. Using RMSSE (Root Mean Square Scaled Error)\cite{georgeathanasopoulos58EvaluatingPoint} as the metric, all sizes of Chronos seem to perform equivalently well, with the best performance observed at a monthly prediction cadence. 

Additionally, in the Ready Tensor analysis Chronos was outperformed at the monthly granularity by a select group of models, including pre-trained NBEATS, VAE, AutoARIMA, and boosting models (XGBoost, Random Forest, Extra Trees, (T)BATS). However, it is important to note that these models require significantly longer inference times when training is considered, highlighting Chronos' efficiency advantage given its pre-trained status. 

The Ready Tensor project also explored the impact of Chronos' parameters on performance. It found that \texttt{'num\_samples'}, \texttt{'top\_p'}, \texttt{'top\_k'}, and \texttt{'temperature'} have minor effects on RMSSE, with only \texttt{'num\_samples'} and \texttt{'top\_k'} showing a just statistically significant 0.05 swing. In terms of compute (CPU, GPU) memory requirements, Chronos fell short of classical models, with only an MLP forecaster on CPU requiring more compute memory for inference. Lastly, the docker image size of Chronos models is at least double that of classical ML methods, a factor to consider if deploying multiple instances; a necessary requirement for multivariate predictions with Chronos. 
 
\section{Future Work}
\label{sec:futurework}

In our data, we have a categorical feature that reflects the weather for the day, normalized temperature, normalized humidity, as well as an \textit{is\_holiday} flag for each day in our dataset. These covariates were not used for the sake of our analysis, however Prophet supports integration of such exogenous variables. Vector Autoregressive Models (VAR) are an extension of ARIMA, allowing for use with multi-variate time-series. They are structured so that each variable is a linear function of itself and past lags of other variables. Furthermore, the python package \textit{Darts} has a Vector Autoregressive Moving Average (VARIMA) implementation which could be used to integrate these covariates to an ARIMA-family forecast. Chronos, as of yet, does not support integration of exogenous time series or other covariates which may increase predictive power for a desired time series, nor does it allow for prediction of multi-variate time series. While we were undertaking our study, Redmond, D. undertook a short study evaluating Chronos' applicability to multi-variate forecasting \cite{redmondArcticbioChronos2024} and concluded that without modification interactions between variables are not adequately preserved, which renders output no more useful than independent and parallel projections. This highlights the need for further development of multivariate models specifically tailored for variable stream analysis.

Now that we have a test bench, we can easily evaluate additional models, including other Chronos sizes. Although this may not generate much useful information, as results from \cite{desaiNewBenchmarkingStudy2024} seem to suggest all sizes of Chronos/T5 perform similarly in an RMSSE sense, akin to our MASE score. Ensemble methods, as suggested by the analysis work in \cite{nixtlaNixtlaExperimentsAmazonchronos} and \cite{statsforecastStatsforecastExperimentsM3} could be a way to further the representational power of Chronos, or guide accuracy among short-term predictions, an area we've shown Chronos struggles. 

An area of further investigation regarding these time-series prediction algorithms is gapped prediction, specifically an evaluation of the effect of a gap in time between the end of context and the beginning of prediction. This would be analogous to needing to anticipate demand for raw materials or components ahead of production; or anticipatory energy demand to ensure generation and distribution planning. Additionally in path planning and navigation, a robot or agent may predict the location of tracked objects before it arrives at a location. 

Although Chronos has demonstrated impressive performance with the T5 model as its foundation, future study may investigate the feasibility and challenges of training alternative models within the Chronos framework. By considering recent advancements in natural language processing; such as \texttt{byT5} \cite{xueByT5TokenfreeFuture2022}, which was trained on byte sequences, rather than SentencePiece tokenization, or \texttt{UL2}, \cite{tayUL2UnifyingLanguage2023} trained using diverse denoising objectives, we can examine the potential benefits and limitations of a Chronos with these underlying models. Furthermore, future research is needed to examine the impact of architectural modifications, such as a Nyströmformer optimized self-attention mechanism \cite{xiongNystromformerNystromBasedAlgorithm2021} or even a reformulation of the network with a Kolmogrov-Arnold Transformer (KAT) \cite{yangKolmogorovArnoldTransformer2024} based on KANs \cite{liuKANKolmogorovArnoldNetworks2024}. 

\section{Conclusion}
\label{sec:conclusion}
This paper presents a comprehensive evaluation of the performance of AWS Chronos and other standard methods for time series forecasting with limited or no tuning. The analysis reveals that Chronos performs especially well in longer prediction tasks, despite its documented limitations. The model's architecture provides resistance to traditional overfitting problems, which is a significant advantage in real-world applications when abundant historical data is available. For very short prediction windows, the superior performance of a naive model suggests that our data is very self-similar at small time horizons, which is consistent with much of industry data. We find that for intermediate forecast horizons (one month) ARIMA strikes the best balance of accuracy when the magnitude of the context does not dwarf the prediction window. Unfortunately Prophet seems very susceptible to overfitting regardless of context. The paper also suggests future work, including the investigation of training alternative models within the Chronos framework.

\newpage

\begin{table*}[p]
\caption{Metrics for $\approx$ 3:1 Calendar Split}
\label{tab:calendar}
\begin{tabular}{ll|llllll}
\toprule
\multicolumn{2}{l}{Time-frame / Series}  & \multicolumn{2}{l}{Week10} & \multicolumn{2}{l}{July} & \multicolumn{2}{l}{Q4} \\ \midrule
Model &  Metric & Casual & Registered & Casual & Registered & Casual & Registered \\
\midrule

\multirow{3}{*}{ARIMA} & EMD & \textbf{0.5211} & 0.5855 & 0.5840 & 1.5559 & 1.0714 & 1.1351 \\
& MASE & 1.4144 & \underline{\textbf{1.0795}} & 1.1360 & 1.5642 & 1.4164 & 1.6249 \\
& WQL & 0.6909 & \textit{\textbf{0.2414}} & 0.4067 & 0.2421 & 0.5212 & 0.2287 \\ 
\cline{1-8}
\multirow{3}{*}{Chronos} & EMD & 1.4943 & 2.1464 & \textbf{0.4580} & 1.0926 & \textbf{0.4498} & \textbf{0.7062} \\
& MASE & 1.4973 & 2.3942 & 1.3188 & 1.1662 & \underline{\textbf{0.8161}} & \underline{ \textbf{1.2207}} \\
& WQL & 0.7158 & 0.4451 & 0.4358 & \textit{\textbf{0.1403}} & \textit{\textbf{0.2653}} & \textit{\textbf{0.1500}} \\ 
\cline{1-8} 
\multirow{3}{*}{Seasonal Naive} & EMD & 0.9529 & \textbf{0.1290} & 0.4904 & \textbf{0.9724} & 1.1879 & 0.7495 \\
& MASE & \underline{\textbf{1.0750}} & 1.0976 & \underline{\textbf{0.8059}} & \underline{\textbf{1.0189}} & 1.3150 & 1.3885 \\
& WQL & \textit{\textbf{0.5251}} & 0.2455 & \textit{\textbf{0.2886}} & 0.1577 & 0.4839 & 0.1954 \\ 
\cline{1-8}
\multirow{3}{*}{Prophet} & EMD & 1.2756 & 0.6780 & 0.7204 & 1.6625 & 1.7304 & 1.6425 \\
& MASE & 1.3100 & 1.5189 & 1.0618 & 1.6668 & 1.8753 & 1.8892 \\
& WQL & 0.5630 & 0.2807 & 0.3035 & 0.2124 & 0.5585 & 0.2285 \\ 
\cline{1-8}
\bottomrule
\end{tabular}
\end{table*}

\newpage

\begin{table*}[p]
\caption{Metrics for ratio 2:1}
\label{tab:two-to-one}
\begin{tabular}{ll|llllll}
\toprule
 \multicolumn{2}{l}{Time-frame / Series}  & \multicolumn{2}{l}{Week10} & \multicolumn{2}{l}{July} & \multicolumn{2}{l}{Q4} \\ \midrule
Model &  Metric & Casual & Registered & Casual & Registered & Casual & Registered \\
\midrule
\multirow{3}{*}{ARIMA} & EMD & 1.3028 & 1.0841 & 1.3181 & 1.7693 & 0.8684 & 1.1875 \\
 & MASE & 1.0631 & \underline{\textbf{0.8559}} & 1.2803 & 1.7088 & 1.1122 & 1.2018 \\
 & WQL & 0.5222 & \textit{\textbf{0.1573}} & 0.4679 & 0.1922 & 0.3065 & \textit{\textbf{0.1345}} \\
\cline{1-8}
\multirow{3}{*}{Chronos} & EMD & 1.4512 & 2.6852 & \textbf{0.7649} & 0.7051 & 0.6409 & 1.1523 \\
 & MASE & 1.4624 & 2.6320 & \underline{\textbf{0.9202}} & \underline{\textbf{1.1081}} & 0.9343 & 1.3134 \\
 & WQL & 0.7566 & 0.5527 & \textit{\textbf{0.4065}} & \textit{\textbf{0.1452}} & 0.2676 & 0.1505 \\
\cline{1-8}
\multirow{3}{*}{Seasonal Naïve} & EMD & \textbf{0.8698} & \textbf{0.1283} & 0.8881 & \textbf{0.7000} & \textbf{0.5313} & \textbf{1.0577} \\
 & MASE & \underline{\textbf{0.9812}} & 1.0916 & 0.9831 & 1.2968 & \underline{\textbf{0.8731}} & \underline{\textbf{1.1083}} \\
 & WQL & 0.5251 & 0.2455 & 0.4839 & 0.1954 & 0.2886 & 0.1577 \\
\cline{1-8}
\multirow{3}{*}{Prophet} & EMD & 1.0894 & 1.0340 & 1.0029 & 1.2323 & 0.6340 & 1.2046 \\
 & MASE & 0.9820 & 0.9731 & 1.1094 & 1.4774 & 0.8815 & 1.2481 \\
 & WQL & \textit{\textbf{0.4212}} & 0.1840 & 0.4362 & 0.1938 & \textit{\textbf{0.2435}} & 0.1389 \\
\cline{1-8}
\bottomrule
\end{tabular}
\end{table*}

\begin{table*}[p]
\caption{Metrics for ratio 3:1}
\label{tab:three-to-one}
\begin{tabular}{ll|llllll}
\toprule
 \multicolumn{2}{l}{Time-frame / Series}  & \multicolumn{2}{l}{Week10} & \multicolumn{2}{l}{July} & \multicolumn{2}{l}{Q4} \\ \midrule
Model &  Metric & Casual & Registered & Casual & Registered & Casual & Registered \\
\midrule
\multirow{3}{*}{ARIMA} & EMD & \textbf{0.8491} & 0.9054 & 1.4505 & 1.3202 & 0.6938 & 1.5873 \\
 & MASE & 1.4083 & \underline{\textbf{1.0777}} & 1.5216 & 1.6230 & 1.1693 & 1.6155 \\
 & WQL & 0.5507 & \textit{\textbf{0.1990}} & 0.4144 & 0.1936 & 0.3414 & 0.1807 \\
\cline{1-8}
\multirow{3}{*}{Chronos} & EMD & 1.5209 & 2.1208 & \textbf{0.7941} & 0.8606 & 0.5010 & 1.1747 \\
 & MASE & 1.5385 & 2.2406 & \underline{\textbf{0.9864}} & \underline{\textbf{1.2700}} & 1.3243 & 1.2893 \\
 & WQL & 0.7254 & 0.4724 & \textit{\textbf{0.3368}} & \textit{\textbf{0.1628}} & 0.4161 & \textit{\textbf{0.1505}} \\
\cline{1-8}
\multirow{3}{*}{Seasonal Naïve} & EMD & 0.9529 & \textbf{0.1290} & 1.1584 & \textbf{0.7200} & \textbf{0.4995} & \textbf{0.9825} \\
 & MASE & \underline{\textbf{1.0750}} & 1.0976 & 1.2822 & 1.3338 & \underline{\textbf{0.8208}} & \underline{\textbf{1.0295}} \\
 & WQL & \textit{\textbf{0.5251}} & 0.2455 & 0.4839 & 0.1954 & \textit{\textbf{0.2886}} & 0.1577 \\
\cline{1-8}
\multirow{3}{*}{Prophet} & EMD & 1.4387 & 0.8653 & 1.4902 & 1.3812 & 0.6979 & 1.4941 \\
 & MASE & 1.4543 & 1.3478 & 1.6716 & 1.6361 & 1.0030 & 1.5694 \\
 & WQL & 0.5849 & 0.2596 & 0.4946 & 0.2043 & 0.2936 & 0.1942 \\
\cline{1-8}
\bottomrule
\end{tabular}
\end{table*}

\begin{table*}[p]
\caption{Metrics for ratio 4:1}
\label{tab:four-to-one}
\begin{tabular}{ll|llllll}
\toprule
 \multicolumn{2}{l}{Time-frame / Series}  & \multicolumn{2}{l}{Week10} & \multicolumn{2}{l}{July} & \multicolumn{2}{l}{Q4} \\ \midrule
Model &  Metric & Casual & Registered & Casual & Registered & Casual & Registered \\
\midrule
\multirow{3}{*}{ARIMA} & EMD & 1.0439 & 0.9837 & 1.5766 & 1.3984 & 0.7779 & 1.9386 \\
 & MASE & 1.2845 & 1.2105 & 1.4349 & 1.6301 & 1.3191 & 2.0174 \\
 & WQL & 0.4444 & 0.1995 & 0.3965 & 0.1812 & 0.3496 & 0.2294 \\
\cline{1-8}
\multirow{3}{*}{Chronos} & EMD & 1.8208 & 1.9466 & \textbf{0.5419} & 0.8308 & \textbf{0.5328} & \textbf{1.0050} \\
 & MASE & 1.8145 & 2.3448 & \underline{\textbf{0.8303}} & \underline{\textbf{1.2224}} & 1.2606 & 1.1344 \\
 & WQL & 0.7449 & 0.3984 & \textit{\textbf{0.2697}} & \textit{\textbf{0.1529}} & 0.3761 & \textit{\textbf{0.1298}} \\
\cline{1-8}
\multirow{3}{*}{Seasonal Naïve} & EMD & 1.1095 & \textbf{0.1452} & 1.1908 & \textbf{0.7505} & 0.5477 & 1.0418 \\
 & MASE & 1.2516 & 1.2352 & 1.3181 & 1.3903 & \underline{\textbf{0.9000}} & \underline{\textbf{1.0916}} \\
 & WQL & 0.5251 & 0.2455 & 0.4839 & 0.1954 & \textit{\textbf{0.2886}} & 0.1577 \\
\cline{1-8}
\multirow{3}{*}{Prophet} & EMD & \textbf{0.9593} & 0.7427 & 1.7236 & 1.7245 & 0.9281 & 2.1993 \\
 & MASE & \underline{\textbf{0.8999}} & \underline{\textbf{1.0282}} & 1.8636 & 1.9740 & 1.1664 & 2.2597 \\
 & WQL & \textit{\textbf{0.3256}} & \textit{\textbf{0.1824}} & 0.5554 & 0.2354 & 0.3165 & 0.2725 \\
\cline{1-8}
\bottomrule
\end{tabular}
\end{table*}

\begin{table*}[p]
\caption{Metrics for ratio 5:1}
\label{tab:five-to-one}
\begin{tabular}{ll|llllll}
\toprule
 \multicolumn{2}{l}{Time-frame / Series}  & \multicolumn{2}{l}{Week10} & \multicolumn{2}{l}{July} & \multicolumn{2}{l}{Q4} \\ \midrule
Model &  Metric & Casual & Registered & Casual & Registered & Casual & Registered \\
\midrule
\multirow{3}{*}{ARIMA} & EMD & 1.0827 & 1.0162 & 1.5273 & 1.4442 & 0.8874 & 2.0370 \\
 & MASE & 1.2411 & 1.2260 & 1.4939 & 1.7562 & 1.4991 & 2.1080 \\
 & WQL & 0.3897 & 0.2002 & 0.4066 & 0.1860 & 0.3425 & 0.2184 \\
\cline{1-8}
\multirow{3}{*}{Chronos} & EMD & 2.1229 & 0.9777 & \textbf{0.5438} & 0.8199 & 0.6773 & \textbf{1.0041} \\
 & MASE & 2.1222 & 1.5000 & \underline{\textbf{0.8161}} & \underline{\textbf{1.2894}} & \underline{\textbf{0.8987}} & \underline{\textbf{1.1110}} \\
 & WQL & 0.7798 & 0.2525 & \textit{\textbf{0.2614}} & \textit{\textbf{0.1497}} & \textit{\textbf{0.2240}} & \textit{\textbf{0.1179}} \\
\cline{1-8}
\multirow{3}{*}{Seasonal Naïve} & EMD & 1.2390 & \textbf{0.1480} & 1.2032 & \textbf{0.8026} & \textbf{0.6398} & 1.1427 \\
 & MASE & 1.3977 & 1.2594 & 1.3319 & 1.4868 & 1.0514 & 1.1973 \\
 & WQL & 0.5251 & 0.2455 & 0.4839 & 0.1954 & 0.2886 & 0.1577 \\
\cline{1-8}
\multirow{3}{*}{Prophet} & EMD & \textbf{1.0656} & 0.7360 & 2.0806 & 1.8680 & 1.1972 & 2.3193 \\
 & MASE & \underline{\textbf{1.0413}} & \underline{\textbf{1.1142}} & 2.2351 & 2.0576 & 1.5350 & 2.3653 \\
 & WQL & \textit{\textbf{0.3459}} & \textit{\textbf{0.1800}} & 0.6665 & 0.2349 & 0.3442 & 0.2640 \\
\cline{1-8}
\bottomrule
\end{tabular}
\end{table*}

\newpage

\bibliographystyle{IEEEbib}
\bibliography{refs}

\end{document}